# FICAug: Feature-Informed Clustering and Augmentation for Facial-Expression-Based Parkinson's Disease Screening

**Yasaman Haghbin**
**Department of Electrical and Computer Engineering, University of Tehran, Iran**
**Hadi Moradi**
**Department of Electrical and Computer Engineering, University of Tehran, Iran**
**Reshad Hosseini**
**Department of Electrical and Computer Engineering, University of Tehran, Iran**

## Abstract
Hypomimia has drawn growing interest as a digital marker for screening Parkinson's disease (PD). However, developing reliable facial-expression-based screening models is challenging because clinical PD datasets are small, exposing models to only a narrow range of how hypomimia can appear across individuals. Standard augmentation strategies do not solve this problem; recombining or perturbing existing samples produces variation, but not new facial configurations that are plausible and clinically meaningful. We introduce FICAug to address this gap. The framework clusters Action Unit (AU) feature vectors extracted from facial expression images, discards clusters that mix labels inconsistently, and generates synthetic AU vectors within the retained clusters with Gaussian sampling. GANimation then reconstructs these synthetic vectors into realistic facial images. A ResNet18 model is pretrained on these reconstructed images, and then fine-tuned on real clinical data. Using the UT-MoDaPark dataset, FICAug achieved 88.63% cross-validation accuracy and 94.00% test accuracy, outperforming both a standard ResNet18 baseline and self-supervised alternatives, including DINO and MSN. These results suggest that synthetic image generation, when guided by label consistency and clinical feature structure, can function as an effective intermediate representation-learning step for facial-expression-based PD screening in settings where clinical data remain scarce.

## Introduction

Parkinson's disease (PD) is a progressive neurodegenerative movement disorder that primarily affects older adults[1]. It is associated with the degeneration of dopaminergic neurons in the substantia nigra, which reduces dopamine production and progressively impairs the control of voluntary movement [2], [3]. Although clinical assessment remains essential for identifying PD, there is growing interest in complementary screening tools that capture disease-related motor characteristics via accessible, non-invasive measurements [4].

The motor manifestations of PD include bradykinesia, resting tremor, muscular stiffness, and rigidity [5]. Bradykinesia can also affect the facial musculature, reducing the speed, intensity, and range of facial movements [6]. This diminished facial expressivity, commonly referred to as hypomimia or facial masking, may be observable during both spontaneous behavior and the deliberate production of emotional expressions [7].

Recent studies have investigated facial expressivity as a digital marker for automatic PD screening. Gomez et al. [8] analyzed facial expressions using both face-image and Action Unit (AU) representations and showed that expression sequences and adaptation to the AU domain can improve discrimination between individuals with PD and healthy controls. More recent studies [9], [10]have modeled temporal derivatives of AU intensities to capture reduced facial movement in early-stage PD and examined the associations among discriminative AUs, hypomimia, and clinical motor scores. These findings demonstrate the relevance of emotion-production tasks, AU configurations, and facial movement dynamics for PD screening. Nevertheless, existing approaches primarily extract or model more discriminative information from the available recordings. They do not directly address the restricted variety of clinically relevant facial configurations represented in small PD datasets.

Developing reliable facial-expression-based screening models, however, is difficult when only small clinical datasets are available. Recruiting patients, collecting standardized recordings, and obtaining reliable labels are resource-intensive processes constrained by ethical, logistical, and population-specific considerations [11], [12], [13]. Models trained directly on such datasets may therefore learn representations that are overly dependent on the limited participants and facial information observed during training.

Related research has approached limited-data representation learning through both feature synthesis and generated images. Feature-space augmentation methods estimate local or class-conditioned feature distributions and sample additional representations to improve few-shot classification [14]. In parallel, StableRep [15] and SynCLR [16] have demonstrated that images produced by generative models can support transferable visual representation learning. Qraitem et al. [17] further showed that separating synthetic-image pretraining from real-image fine-tuning can reduce reliance on artifacts or correlations introduced by mixing the two domains during training. However, feature-space methods generally use the synthesized representations directly for classification, whereas synthetic-image pretraining methods typically generate images from text descriptions or class names using large pretrained models. Consequently, neither direction explicitly uses label-consistent, task-specific clinical features to construct an intermediate image domain for learning representations relevant to PD screening.

The central challenge is thus not only the limited number of training data, but also the restricted representational coverage of the available data. A small dataset exposes a model to only a narrow subset of how hypomimia may appear across individuals, emotional expressions, and levels of motor impairment. Conventional image augmentation techniques, such as geometric transformations, Gaussian perturbations, cutout, and Mixup [18], [19], [20], can improve invariance to predefined transformations. Nevertheless, they primarily modify or combine existing observations and may not introduce task-relevant facial configurations that are plausible in the underlying feature space. This limitation can prevent the model from developing expressive representations of the subtle patterns associated with PD.

To address this problem, we introduce the FICAug (Feature-Informed Clustering and Augmentation) framework. The proposed framework clusters the extracted features and evaluates each cluster for label purity. Only clusters with unambiguous class association are retained for synthesis. Within these clusters, synthetic features are generated via Gaussian sampling and subsequently mapped back into realistic facial images using a generative adversarial network. A convolutional neural network (CNN) is first trained on the reconstructed images and subsequently fine-tuned on the original dataset. FICAug uses synthetic samples as an intermediate learning environment in which a convolutional neural network can encounter label-consistent facial configurations that were not directly present in the original image set.

## Method

Figure 1 presents an overview of the proposed FICAug framework.

### Dataset and Facial Representation

The Facial Action Coding System (FACS) is an anatomically based framework for describing observable facial movements. FACS decomposes a facial expression into individual muscle-related components known as Action Units (AUs) [21]. Changes in the activation and intensity of these AUs have been used to characterize hypomimia and support PD screening [7].

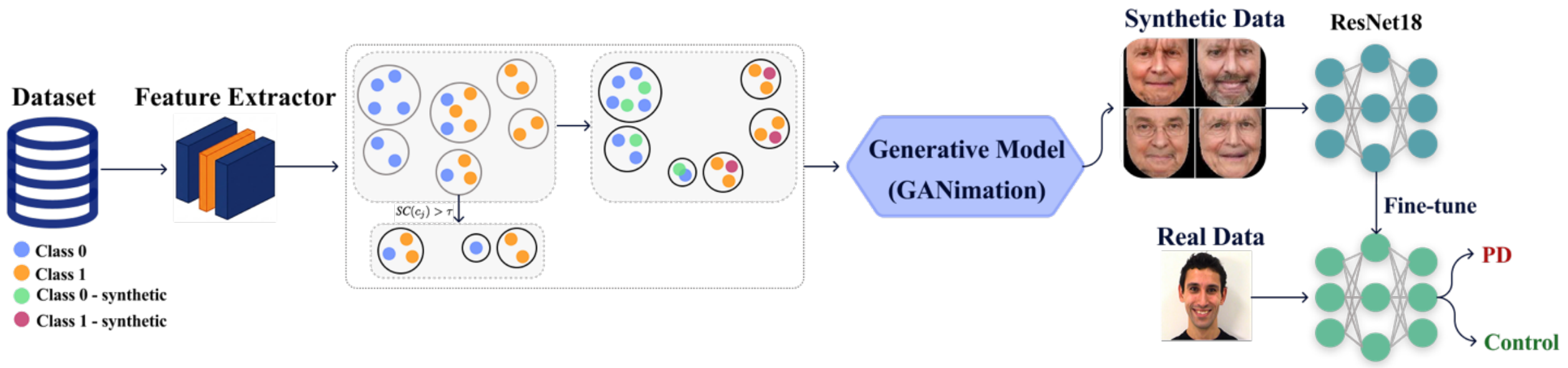

**Figure 1. Overview of the proposed FICAug framework. Features are extracted from the dataset and clustered to identify label-consistent regions. Synthetic feature vectors are sampled from these regions and reconstructed as facial images using GANimation. A ResNet18 model is first trained on the reconstructed images and subsequently fine-tuned on the original data for Parkinson's disease (PD) versus control classification.**

We used the UT-MoDaPark dataset [22], comprising videos from 55 individuals captured using an Android tablet camera. Before each recording, the participant watched a short demonstration of the required task. The participant was then instructed to produce the corresponding facial expression and maintain it for several seconds. Five expressions were used in this study: anger, happiness, disgust, fear, and surprise. Samples are shown in Figure 2.

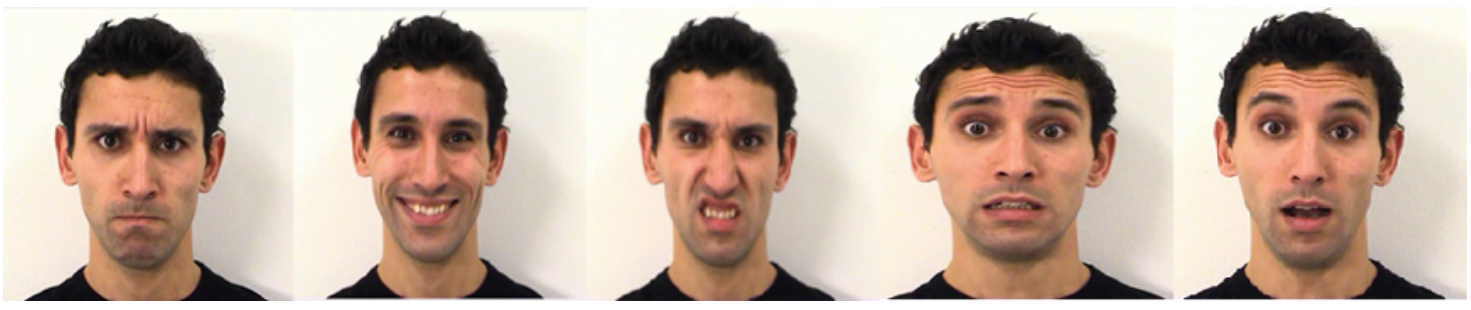

**Figure 2. Facial expressions from left to right: anger, happiness, disgust, fear, and surprise**

Each video captures the transition from a neutral facial configuration to the intended emotional expression. To obtain a consistent image-level representation, the frame corresponding to the peak of the expression was selected from each video and used in the subsequent analysis.
We used OpenFace [23] to extract AU from each image. Each AU corresponds to the movement of a specific group of facial muscles. OpenFace reports both a binary activation value and a continuous intensity value ranging from 0 to 5 for each detected AU. Three AUs were selected for each emotional category. The selection was based on the mean AU values observed for the corresponding expression, the known association between each AU and the facial movement involved in the expression, and Pearson correlation analysis. The selected AUs are listed in Table 1.

**Table 1. Associated AUs to each facial expression**

| Emotion | Facial Action Unit |
|---|---|
| Rage | AU09 (Nose Wrinkler) |
| | AU10 (Upper Lip Raiser) |
| | AU20 (Lip Stretcher) |
| Laugh | AU15 (Lip Corner Depressor) |
| | AU17 (Chin Raiser) |
| | AU25 (Lips Part) |
| Disgust | AU23 (Lip Tightener) |
| | AU25 (Lips Part) |
| | AU45 (Blink) |
| Fear | AU05 (Upper Lid Raiser) |
| | AU12 (Lip Corner Puller) |
| | AU17 (Chin Raiser) |
| Surprise | AU04 (Brow Lowerer) |
| | AU05 (Upper Lid Raiser) |
| | AU15 (Lip Corner Depressor) |

For each expression, the selected AU intensities form a feature vector. Let

$$X = \{x_1, x_2, \dots, x_n\}, \qquad x_i \in \mathbb{R}^d, \tag{1}$$

denote the resulting AU feature set, where ($d$ =3) is the number of selected AUs for the corresponding expression. Each feature vector ($x_i$) is associated with a class label ($y_i$), indicating whether the sample belongs to the PD group or the control group. The subsequent clustering and synthesis procedure is applied to these AU representations.

**Clustering**

K-means clustering is applied to the AU feature vectors to obtain $C = c_1, c_2, \dots, c_k$. The K-means algorithm operates iteratively to minimize the total sum of distances from each point in a cluster to its centroid [24]. Each cluster $c_j$ comprises data points $x_{i,j}$ where $i = 1,2,\dots,n_j$, with $n_j$ representing the number of data points in cluster $c_j$.

The number of clusters $k$ is a hyperparameter of FICAug and is set to exceed the number of classification labels. An initial value can be determined using a clustering-selection procedure such as the elbow method [25].

**Cluster Evaluation and Recursive Re-Clustering**

After the initial clustering stage, each cluster is evaluated according to the labels of its members. A cluster containing samples from only one class is considered label-consistent and is retained for synthetic feature generation. Its synthetic samples inherit the PD or control label represented by the cluster.

A cluster containing both PD and control samples is considered mixed. Synthetic sampling is not performed directly within such a cluster because doing so could generate feature vectors in ambiguous regions between the two classes. Instead, a separation criterion is used to determine whether the mixed cluster contains class-specific substructure that can be recovered through further clustering.

For a mixed cluster ($c_j$), let ($l_j$) denote the set of class labels represented in that cluster. The Class Separation Metric (CSM) is defined as the ratio of inter-class separation to intra-class cohesion:

$$\text{CSM}(c_j) = \frac{\mathcal{S}_{inter}(c_j)}{\mathcal{C}_{intra}(c_j)} \tag{2}$$

where,

$$\mathcal{C}_{intra}(c_j) = \sum_{l \in \mathcal{L}} \frac{1}{n_l(n_l - 1)} \sum_{\substack{i,k \in l \\ i \neq k}} \| x_i - x_k \| \tag{3}$$

$$\mathcal{S}_{inter}(c_j) = \sum_{\substack{l_i, l_k \in \mathcal{L} \\ l_i \neq l_j}} \frac{1}{n_{l_i} \cdot n_{l_k}} \sum_{\substack{x_i \in l_i \\ x_k \in l_k}} \| x_i - x_k \| \tag{4}$$

$\mathcal{C}_{intra}$ measures the average pairwise distance between all points with the same label within a cluster, where $n_l$ is the number of points in class l, and $\|x_i - x_k\|$ is the Euclidean distance between points within the same class. $\mathcal{S}_{inter}$ measures the average pairwise distance between points in different classes $l_i$ and $l_k$, where $n_{l_i}$ and $n_{l_k}$ are the numbers of points in classes $l_i$ and $l_k$, respectively.

This term measures the average pairwise distance between samples with different class labels. A high CSM value indicates that samples from different classes are relatively far apart in comparison with the distances among samples of the same class.

The label composition of the cluster is additionally characterized using entropy:

$$H(c_j) = -\sum_{l \in \mathcal{L}_j} p_j\,(l) \log_2 p_j(l), \tag{5}$$

where $p_j(l)$ is the proportion of samples with label $l$ in cluster $c_j$. A lower entropy indicates that the cluster is dominated by one class, whereas a higher entropy indicates a more balanced mixture of PD and control samples.

Finally, the two quantities are combined as

$$\text{SC}(c_j) = \frac{\text{CSM}(c_j)}{H(c_j)}, \tag{6}$$

where $SC(c_j)$ denotes the separation criterion. The resulting values are normalized to the interval [0,1].

For a mixed cluster, a high normalized separation criterion indicates that its class-specific samples possess sufficient geometric separation to justify another clustering step. Therefore, if

$$\text{SC}(c_j) > \tau, \tag{7}$$

where (τ) is a predefined threshold. During this local re-clustering step, the number of subclusters is set to the number of classes in the screening task. Each resulting subcluster is then evaluated using the same procedure.
If a mixed cluster does not satisfy the threshold, it is not used for synthetic sampling. The recursive procedure therefore produces a set of retained clusters that are label-consistent while excluding unresolved regions of class overlap.
The threshold ($\tau$) controls the extent of recursive subdivision. A value closer to one permits fewer mixed clusters to undergo re-clustering, resulting in a simpler partition. In contrast, a value closer to zero permits more mixed clusters to be subdivided and can produce a more detailed partition of the feature space. At the limiting cases, ($\tau$=1) effectively restricts the procedure to the initial K-means partition, whereas ($\tau$=0) permits recursive subdivision to continue until label-consistent clusters are obtained or no suitable subdivision remains.

**Cluster-Guided Feature Synthesis**
Clusters that contain data points from a single class are suitable for generating synthetic data. On the other hand, clusters that contain data points from multiple classes after assessing the separation criterion are considered mixed and are typically discarded. We produce synthetic data points based on the estimated parameters of clusters containing points from a particular class using a normal distribution. It depicts the process of obtaining new artificial data points starting with the clustering radius and the center to fit the observed distribution.
For clusters with more than one data point, the cluster radius $r_j$ is calculated based on the maximum distance between any point in the cluster and the cluster center $\mu_j$, plus a small augmentation factor proportional to the range of these distances. This ensures that the radius takes into account the spread of points around the center, while also allowing for a slight margin beyond the maximum distance. For clusters with only a single data point, the radius is set to a constant value, which is derived from the average standard deviation of the features, scaled by a small factor. Therefore, the radius $r_j$ for each cluster $j$ is defined as:

$$r_j \leftarrow \begin{cases} 0.01 \times \frac{1}{d}\sum_{i=1}^{d} \sigma_j & \text{if } n_{c_j} == 1 \\ \max\left(D_{x,\mu_j}\right) + 0.1 \times \left(\max\left(D_{x,\mu_j}\right) - \min\left(D_{x,\mu_j}\right)\right) & \text{if } n_{c_j} > 1 \end{cases} \quad (8)$$

$\sigma_j$ is the standard deviation of the $j$-th feature calculated based on the data points from the previous clustering step. This indicates that the variability is measured considering the feature distribution observed in the last iteration or phase of clustering. Also, $D_{x,\mu_j}$ refers to the distance of each point $x \in X_j$ from the cluster center $\mu_j$.
The generation of synthetic data points involves sampling $q$ artificial data points from a normal distribution centered at the cluster mean $\mu_j$, as depicted in Equation 9:

$$x_{new,q} \sim N\left(\mu_j, \Sigma_j\right) \quad (9)$$

$\Sigma_j$ represents the covariance matrix calculated based on the variance of $r_j$.
The number of synthetic data points generated in each cluster is proportional to $n_{c_j}$, denoted by the coefficient $\alpha$.

$$q = \alpha \cdot n_{c_j} \quad (10)$$

The coefficient $\alpha$ controls the number of synthetic data points against the number of data points of the given set. It is flexible according to the requirements of a specific dataset and clustering needs. This proportional augmentation approach makes it possible to assign more synthetic data points to areas of the input space that are densely occupied by training data points.

**Feature-to-Image Reconstruction**
The synthetic vectors generated in the AU feature space describe facial-muscle activation patterns rather than complete facial images. To use these vectors for visual representation learning, FICAug projects them back into the image domain using an AU-conditioned facial-expression generation model.
To initiate the generation process in the original image space, we first utilized the Generated Photos tool [26] to create a collection of neutral facial images. This tool employs GAN-based models to synthesize realistic human faces. To increase the realism and relevance of the dataset, we specifically selected images of elderly individuals with neutral expressions (i.e., without any emotional cues). These images provided a strong foundation for applying the synthetic AUs. Examples of these synthesized neutral faces are shown in Figure 3.

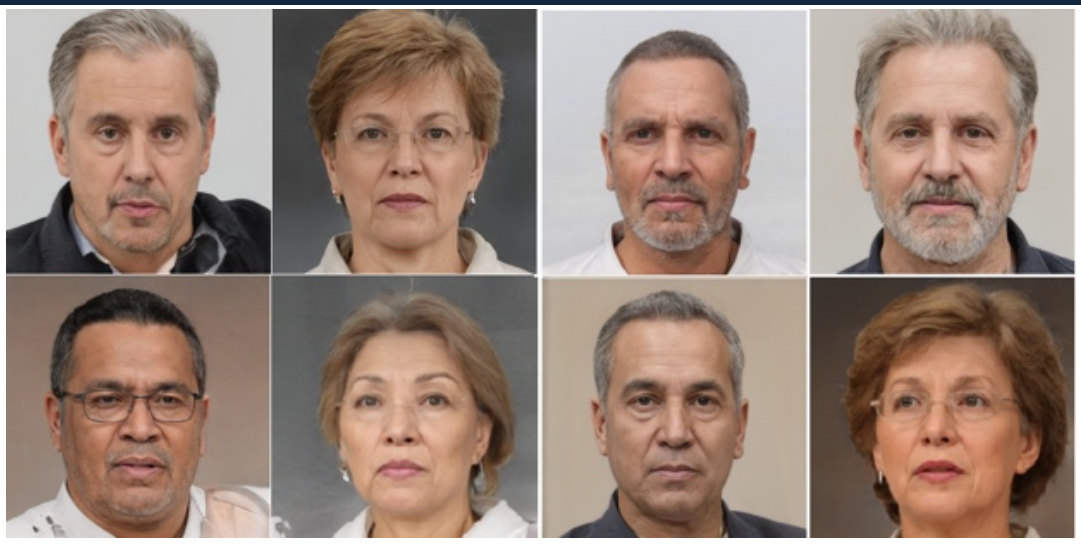

**Figure 3. AI-generated neutral face images used for applying Action Units**

GANimation [27] is used to reconstruct the synthetic AU vectors as facial expressions. GANimation is an AU-conditioned generative adversarial network that supports continuous control over facial movements without requiring paired images. Its attention and deformation mechanisms modify the facial regions associated with the supplied AU values while preserving the remaining visual characteristics of the input face.
The GANimation model was fine-tuned for 20 epochs using the selected neutral face images. This step adapted the generator to the visual characteristics of the neutral-face set used in the reconstruction procedure.
For each emotional category, the conditioning information supplied to GANimation includes the dominant emotion-related AUs selected in Table 1. Their intensity values are provided by the cluster-guided synthetic vectors generated in Equation 9. Additional AU values are sampled from a normal distribution to introduce further expression variation, following the same generation procedure used in the original implementation.
GANimation applies each AU conditioning vector to a neutral face and produces a reconstructed emotional-expression image. The class label assigned to the reconstructed image is inherited from the label-consistent cluster from which the AU vector was sampled.
The same set of neutral identities is used to generate images for the PD and control classes. Therefore, the primary systematic difference between the reconstructed class-specific images is the AU conditioning pattern rather than the identity of the neutral face. This design encourages the subsequent CNN to learn differences in facial movement and expression rather than relying on unrelated identity characteristics. Examples of reconstructed control and Parkinson's disease expressions are shown in Figures 4 and 5, respectively.

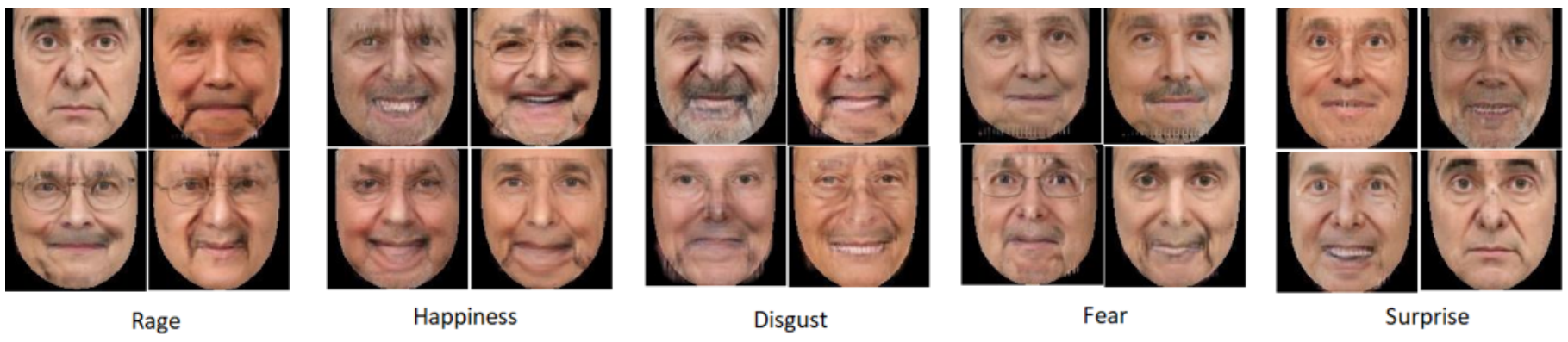

**Figure 4. Synthetic facial expressions generated for control individuals using GANimation**

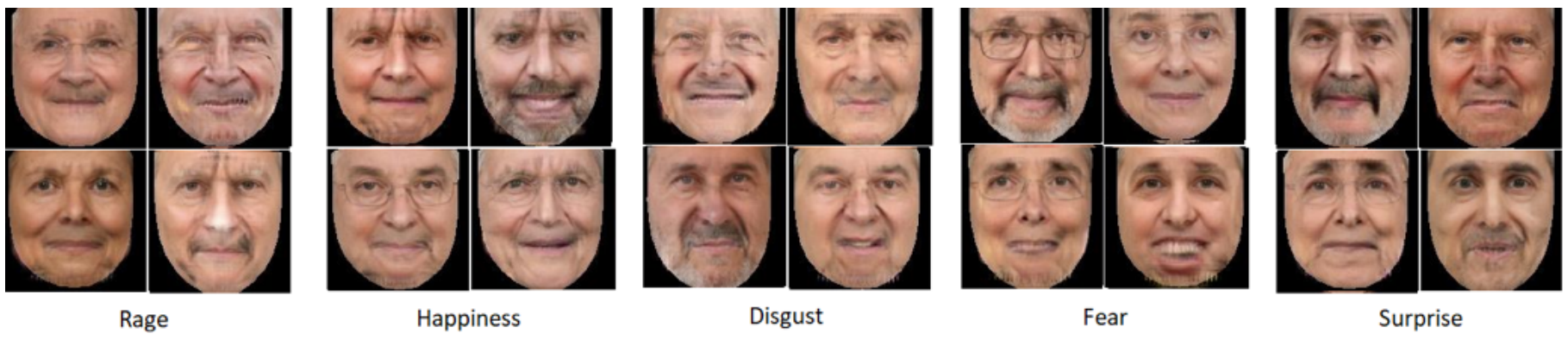

**Figure 5. Synthetic facial expressions generated for Parkinson's patients using GANimation**

## CNN Model Training and Fine-Tuning

The reconstructed facial images are used to bootstrap the representation learned by the CNN. In the first training stage, the ResNet18 [28] is trained exclusively on the reconstructed images and their inherited PD or control labels. Because these images encode structured AU variations sampled from label-consistent regions, this stage exposes the network to facial-movement configurations that are plausible under the original feature distribution but are not directly present in the clinical image set.
The purpose of this stage is not merely to increase the number of training images. Instead, it provides an intermediate representation-learning task in which the ResNet18 learns to distinguish class-associated patterns of facial movement across controlled variations of expression and neutral-face identity.
After the initial training stage, the ResNet18 is fine-tuned on the original images from the UT-MoDaPark dataset. This fine-tuned network performs the PD screening task by classifying each real facial-expression image as belonging to the PD or control group.

## Experiments

In this section, we evaluate the effectiveness of the proposed FICAug framework for facial-expression-based PD screening. Rather than treating synthetic sample generation as the final objective, the evaluation examines the role of each stage in supporting the final image-based screening model.

Specifically, the experiments address three questions. First, do the cluster-guided synthetic AU vectors preserve the statistical and discriminative structure of the original feature space? Second, can the reconstruction model reproduce the intended AU configurations in the generated facial images? Finally, does pretraining on these reconstructed images improve PD screening after fine-tuning on real data?

### Data Splitting for Training, Validation, and Testing

We used a leave-two-subjects-out cross-validation procedure on the training and validation set. The procedure produced 25 folds, with one participant with PD and one control participant held out for validation in each fold. The remaining participants were used for training. This subject-level separation prevents images from the same validation participant from appearing in the corresponding training set.

**Table 2. Distribution of training, validation and test sets for each emotion**

| Dataset | Group | Anger | Happiness | Disgust | Fear | Surprise |
|---|---|---|---|---|---|---|
| Train/Val | Parkinson | 22 | 18 | 23 | 22 | 22 |
| | Control | 19 | 19 | 17 | 19 | 17 |
| Test | Parkinson | 4 | 7 | 3 | 2 | 7 |
| | Control | 9 | 11 | 11 | 11 | 9 |

The independent test set was not used during model fitting or hyperparameter selection. Table 2 reports the numbers of PD and control samples available for each facial-expression task in the training/validation and test partitions. The partitions are balanced in age and gender distributions between the two groups.

For the aggregated evaluation across facial expressions, predictions from the 25 cross-validation models were combined through weighted majority voting. The contribution of each model was weighted according to its validation performance.

### Evaluation of the AU Feature Space

The first group of experiments evaluates the AU vectors produced by the cluster-guided synthesis procedure. These experiments are not intended to establish feature-space classification as the primary output of FICAug. Instead, they examine whether the synthesized vectors provide a credible and label-informative foundation for the subsequent image-reconstruction and representation-learning stages.

**Table 3. Statistical comparison between features of real and synthetic data using t-test, Levene's test, and Kolmogorov–Smirnov (KS) test, with corresponding p-values.**

| | | t-test | | Levene-test | | ks-test | |
|---|---|---|---|---|---|---|---|
| **Emotion** | **Feature** | **Control** | **Parkinson** | **Control** | **Parkinson** | **Control** | **Parkinson** |
| Rage | AU09 | 0.77 | 0.72 | 0.71 | 0.71 | 0.33 | 0.83 |
| | AU10 | 0.83 | 0.10 | 0.97 | 0.78 | 0.83 | 0.17 |
| | AU20 | 0.36 | 0.79 | 0.71 | 0.34 | 0.57 | 0.57 |
| Happiness | AU15 | 0.21 | 0.64 | 0.23 | 0.72 | 0.01 | 0.98 |
| | AU17 | 0.01 | 0.53 | 0.01 | 0.94 | 0.01 | 0.57 |
| | AU25 | 0.22 | 0.82 | 0.47 | 0.32 | 0.33 | 0.83 |
| Disgust | AU23 | 0.36 | 0.48 | 0.32 | 0.54 | 0.33 | 0.83 |
| | AU25 | 0.14 | 0.30 | 0.08 | 0.83 | 0.17 | 0.33 |
| | AU45 | 0.07 | 0.99 | 0.13 | 0.46 | 0.57 | 0.98 |
| Fear | AU05 | 0.87 | 0.27 | 0.85 | 0.05 | 0.98 | 0.33 |
| | AU12 | 0.46 | 0.72 | 0.45 | 0.24 | 0.17 | 0.57 |
| | AU17 | 0.31 | 0.16 | 0.31 | 0.52 | 0.98 | 0.57 |
| Surprise | AU04 | 0.92 | 0.35 | 0.92 | 0.39 | 0.33 | 0.83 |
| | AU05 | 0.19 | 0.70 | 0.54 | 0.65 | 0.33 | 0.83 |
| | AU15 | 0.85 | 0.15 | 0.26 | 0.77 | 0.17 | 0.33 |

We first examined whether the synthesized AU vectors broadly preserved the statistical characteristics of the original AU observations. For each facial-expression category and clinical group, 20 samples were randomly selected from the real and synthetic feature sets.

Three statistical tests were applied independently to each selected AU. A t-test [29] was used to compare the feature means, Levene's test [30] was used to compare the variances, and the Kolmogorov-Smirnov test [31] was used to compare the empirical distributions. The resulting p-values are reported in Table 3.

Using ($p<0.05$) as the rejection criterion, 86 of the 90 reported comparisons did not indicate a statistically detectable difference between the corresponding real and synthetic feature distributions. Four comparisons produced p-values below 0.05, all involving control samples for the happiness expression. These results indicate that the cluster-guided synthesis procedure generally retains the first-order statistical structure of the original AU observations while introducing controlled variation within the selected label-consistent regions.

We next evaluated whether the cluster-guided AU vectors retained information relevant to distinguishing the PD and control groups. This analysis serves as an intermediate validation of the feature-space regions used to construct the reconstructed-image training set.

We compared FICAug against three approaches: a baseline (no augmentation), Gaussian noise (GN) perturbation, and the kNNMTD algorithm [32], using four classifiers across five emotions. The comparison was conducted using four classifiers—K-nearest neighbors (KNN), support vector machine (SVM), random forest (RF), and multilayer perceptron (MLP)—across five facial expressions.

**Table 4. Comparison between accuracy of FICAug, GN, kNNMTD and Baseline on 5 different emotions.**

| | | KNN | | SVM | | RF | | MLP | |
|---|---|---|---|---|---|---|---|---|---|
| **Emotion** | **Method** | **Validation** | **Test** | **Validation** | **Test** | **Validation** | **Test** | **Validation** | **Test** |
| Rage | Baseline | 54.00 | 69.23 | 52.00 | 53.84 | 64.00 | 46.15 | 54.00 | 76.92 |
| | GN | 58.00 | 69.23 | 56.00 | 53.84 | 64.00 | 46.15 | 64.00 | 76.92 |
| | kNNMTD | 54.00 | 69.23 | 56.00 | 53.84 | 64.00 | 53.84 | 54.00 | 69.23 |
| | FICAug | 60.00 | 76.92 | 58.00 | 69.23 | 68.00 | 53.84 | 64.00 | 84.61 |
| Happiness | Baseline | 70.83 | 61.11 | 70.83 | 72.22 | 79.17 | 66.66 | 75.00 | 77.77 |
| | GN | 70.83 | 61.11 | 70.83 | 72.22 | 79.17 | 66.66 | 79.17 | 66.66 |
| | kNNMTD | 70.83 | 61.11 | 70.83 | 72.22 | 79.17 | 66.66 | 68.75 | 72.22 |
| | FICAug | 70.83 | 61.11 | 79.17 | 83.33 | 81.25 | 72.22 | 81.25 | 77.77 |
| Disgust | Baseline | 68.00 | 42.85 | 76.00 | 42.85 | 78.00 | 57.14 | 70.00 | 57.14 |
| | GN | 76.00 | 50.00 | 80.00 | 35.71 | 82.00 | 64.28 | 80.00 | 35.71 |
| | kNNMTD | 68.00 | 42.85 | 76.00 | 42.85 | 84.00 | 57.14 | 68.00 | 50.0 |
| | FICAug | 72.00 | 64.28 | 82.00 | 50.00 | 84.00 | 64.28 | 82.00 | 64.28 |
| Fear | Baseline | 60.00 | 69.23 | 68.00 | 84.62 | 58.00 | 69.23 | 62.00 | 61.53 |
| | GN | 64.00 | 69.23 | 68.00 | 76.92 | 64.00 | 69.23 | 70.00 | 69.23 |
| | kNNMTD | 60.00 | 69.23 | 68.00 | 76.92 | 72.00 | 53.84 | 72.00 | 61.53 |
| | FICAug | 62.00 | 69.23 | 72.00 | 76.92 | 74.00 | 69.23 | 76.00 | 69.23 |
| Surprise | Baseline | 68.00 | 56.25 | 78.00 | 43.75 | 70.00 | 43.75 | 72.00 | 62.50 |
| | GN | 68.00 | 50.00 | 70.00 | 43.75 | 80.00 | 50.00 | 78.00 | 43.75 |
| | kNNMTD | 68.00 | 56.25 | 68.00 | 50.00 | 68.00 | 50.00 | 74.00 | 62.50 |
| | FICAug | 70.00 | 62.50 | 78.00 | 50.00 | 82.00 | 56.25 | 82.00 | 75.00 |
| All Emotions | Baseline | 68.18 | 61.11 | 72.72 | 66.66 | 75.00 | 66.66 | 68.18 | 72.22 |
| | GN | 70.45 | 66.66 | 72.72 | 61.11 | 82.36 | 66.66 | 81.81 | 66.66 |
| | kNNMTD | 68.18 | 61.11 | 77.27 | 66.66 | 79.54 | 66.66 | 75.00 | 77.77 |
| | FICAug | 75.00 | 72.22 | 79.54 | 72.22 | 84.09 | 77.77 | 81.81 | 83.33 |

As shown in Table 4, for the anger expression, FICAug obtained the highest test accuracy with three of the four classifiers. Its strongest result was 84.61% with the MLP, compared with 76.92% for both the unaugmented baseline and Gaussian-noise augmentation. For happiness, FICAug reached a test accuracy of 83.33% with the SVM and improved the RF result from 66.66% to 72.22%.

For disgust, FICAug obtained a test accuracy of 64.28% with KNN, RF, and the MLP. The RF result was tied with Gaussian-noise augmentation, whereas the k-nearest-neighbor and MLP results exceeded the corresponding alternatives. For fear, FICAug improved the validation results but did not consistently outperform all alternatives on the test set. In particular, the SVM trained without augmentation obtained the highest test accuracy of 84.62%. For surprise, FICAug produced the strongest MLP result, reaching 82.00% validation accuracy and 75.00% test accuracy. When the predictions from all facial expressions were aggregated, FICAug produced the highest test accuracy for every classifier. It reached validation and test accuracies of 75.00% and 72.22% with KNN, 79.54% and 72.22% with the SVM, 84.09% and 77.77% with the RF, and 81.81% and 83.33% with the MLP.

The feature-space results show that the retained clusters contain AU configurations that remain useful for PD discrimination after sampling. However, performance at this stage is not the main objective of FICAug.

**Fidelity of Feature-to-Image Reconstruction**

The next experiment evaluates whether GANimation transfers the synthesized AU vectors into the reconstructed images with acceptable fidelity. For each reconstructed image, OpenFace was used to re-extract the AU intensities. The extracted values were compared with the corresponding synthetic AU values supplied to GANimation. Table 5 presents the MAE values for each emotion and both subject groups.

The errors ranged from 0.05 to 0.19. For the PD group, the lowest error was observed for disgust at 0.08, while the highest was observed for fear at 0.19. For the control group, the lowest error was for happiness at 0.05, and the highest for anger at 0.16. These results indicate that the reconstructed images generally preserve the intended AU conditioning values.

**Table 5. Evaluation of GANimation using MAE**

| **Emotion** | **Rage** | **Happiness** | **Disgust** | **Fear** | **Surprise** |
|---|---|---|---|---|---|
| Parkinson's | 0.17 | 0.09 | 0.08 | 0.19 | 0.09 |
| Control | 0.16 | 0.05 | 0.07 | 0.06 | 0.07 |

**Evaluation of CNN Representation Bootstrapping**

The primary experiment evaluates whether pretraining on the reconstructed facial images improves the representation used for PD screening. Table 6 presents the classification results of FICAug compared to three baselines: a standard ResNet18, a DINO-pretrained ResNet50 [33], and the Masked Siamese Networks (MSN) [34] model using *facebook/vit-msn-small*. All models used the same subject partitions and downstream evaluation protocol.

**Table 6. Comparison of classification accuracy between FICAug and other algorithms**

| **Emotion** | **Method** | **Cross Validation** | **Test** |
|---|---|---|---|
| Rage | Baseline | 62.00 | 53.84 |
| | DINO | 44.00 | 30.76 |
| | MSN | 70.0 | 61.53 |
| | **FICAug** | **74.00** | **69.23** |
| Happiness | Baseline | 72.91 | 72.22 |
| | DINO | 60.41 | 55.55 |
| | MSN | **83.33** | 55.55 |
| | **FICAug** | 81.25 | **77.77** |
| Disgust | Baseline | 82.00 | 71.42 |
| | DINO | 62.00 | 50.00 |
| | MSN | 82.00 | 64.28 |
| | **FICAug** | **86.00** | **78.57** |
| Fear | Baseline | 70.00 | 69.23 |
| | DINO | 52.00 | 30.76 |
| | MSN | 76.00 | 69.23 |
| | **FICAug** | **78.00** | **76.92** |
| Surprise | Baseline | 68.00 | 75.00 |
| | DINO | 40.00 | 56.25 |
| | MSN | **80.00** | 62.50 |
| | **FICAug** | 68.00 | **75.00** |
| All Emotions | Baseline | 70.96 | 72.22 |
| | DINO | 50.00 | 66.66 |
| | MSN | 68.18 | 72.22 |
| | **FICAug** | **88.63** | **94.00** |

For anger, FICAug obtained a test accuracy of 69.23%, exceeding the standard ResNet18 result of 53.84%, the DINO result of 30.76%, and the MSN result of 61.53%. For happiness, MSN obtained the highest cross-validation accuracy at 83.33%, but its test accuracy was 55.55%. FICAug obtained 81.25% cross-validation accuracy and 77.77% test accuracy.

For disgust, FICAug achieved the highest cross-validation and test accuracies, reaching 86.00% and 78.57%, respectively. For fear, it similarly produced the highest results, with 78.00% cross-validation accuracy and 76.92% test accuracy. For surprise, FICAug matched the standard ResNet18 test accuracy of 75.00%, while DINO and MSN obtained 56.25% and 62.50%, respectively.

The clearest difference was observed after aggregating predictions across all facial expressions. FICAug achieved a cross-validation accuracy of 88.63% and a test accuracy of 94.00%. The standard ResNet18 obtained 70.96% and 72.22%, DINO obtained 50.00% and 66.66%, and MSN obtained 68.18% and 72.22%. These results show that the feature-to-image training sequence provides a stronger initialization for the final PD screening task than the evaluated image-based baselines.

## Conclusion

This paper introduced FICAug, a framework for facial-expression-based Parkinson's disease (PD) screening that addresses the limited representational coverage of small clinical datasets. Rather than applying conventional image augmentation, FICAug clusters Action Unit (AU) feature vectors, restricts synthesis to clusters with unambiguous class association, and projects the resulting synthetic vectors back into the image domain using an AU-conditioned generative model. This synthetic image set is then used to bootstrap a convolutional neural network (CNN) before fine-tuning on real clinical recordings.

Using UT-MoDaPark dataset, FICAug achieved a cross-validation accuracy of 88.63% and a test accuracy of 94.00%, compared with 70.96% and 72.22%, respectively, for the standard ResNet18 baseline. These results support the central hypothesis of this study: information already present in a limited clinical feature space can be selectively explored, reconstructed into new image-domain learning examples, and transferred back to the real Parkinson's disease screening task through fine-tuning.

The study has several limitations. The UT-MoDaPark dataset, while carefully partitioned at the subject level, remains modest in size. Additionally, GANimation's reconstruction fidelity, while acceptable on average, varied across expressions and groups, and further improvements in AU-conditioned generation could enhance the realism and label consistency of the synthetic training images. Future work could explore extending FICAug to a video-based dataset. Overall, these results support the broader premise that label-consistent, feature-guided synthetic image generation offers a viable strategy for improving facial-expression-based PD screening when clinical data are scarce, and they motivate further exploration of intermediate synthetic representation learning for other movement-disorder screening tasks constrained by similarly limited data availability.